\documentclass[11pt,a4paper]{article}
\usepackage[hyperref]{acl2018}
\usepackage{times}
\usepackage{latexsym}
\usepackage{graphicx}
\usepackage{amsmath}
\usepackage{amsfonts}
\usepackage{url}
\usepackage{multirow,tabularx}

\usepackage{url}

\aclfinalcopy 

\title{Sentence Modeling via Multiple Word Embeddings and \\ Multi-level Comparison for Semantic Textual Similarity}

\author{Nguyen Huy Tien\textsuperscript{1}, Nguyen Minh Le\textsuperscript{1}, Yamasaki Tomohiro\textsuperscript{2}, Izuha Tatsuya\textsuperscript{2}
		\\ \textsuperscript{1}Japan Advanced Institute of Science and Technology (JAIST) 
        \\ \textsuperscript{2}Toshiba Research \& Development Center, Japan
        \\ \tt ntienhuy@jaist.ac.jp, nguyenml@jaist.ac.jp, 
        \\ \tt tomohiro2.yamasaki@toshiba.co.jp, tatsuya.izuha@toshiba.co.jp }

\date{}

\begin{document}
\maketitle
\begin{abstract}
Different word embedding models capture different aspects of linguistic properties. This inspired us to propose a model (M-MaxLSTM-CNN) for employing multiple sets of word embeddings for evaluating sentence similarity/relation. Representing each word by multiple word embeddings, the MaxLSTM-CNN encoder generates a novel sentence embedding. We then learn the similarity/relation between our sentence embeddings via Multi-level comparison. Our method M-MaxLSTM-CNN consistently shows strong performances in several tasks (i.e., measure textual similarity, identify paraphrase, recognize textual entailment). According to the experimental results on STS Benchmark dataset and SICK dataset from SemEval, M-MaxLSTM-CNN outperforms the state-of-the-art methods for textual similarity tasks. Our model does not use hand-crafted features (e.g., alignment features, Ngram overlaps, dependency features) as well as does not require pre-trained word embeddings to have the same dimension.
\end{abstract}

\section{Introduction}
Measuring the semantic similarity/relation of two pieces of short text plays a fundamental role in a variety of language processing tasks (i.e., plagiarism detection, question answering, and machine translation). Semantic textual similarity (STS) task is challenging because of the diversity of linguistic expression. For example, two sentences with different lexicons could have a similar meaning. Moreover, the task requires to measure similarity at several levels (e.g., word level, phrase level, sentence level). These challenges give difficulties to conventional approaches using hand-crafted features. 

Recently, the emergence of word embedding techniques, which encode the semantic properties of a word into a low dimension vector, leads to the successes of many learning models in natural language processing (NLP). For example, \citet{Kalchbrenner} randomly initialize word vectors, then tunes them during the training phase of a sentence classification task. By contrast, \citet{huy} initialize word vectors via the pre-train word2vec model trained on Google News \citep{mikolov2013distributed}. \citet{wieting2015ppdb} train a word embedding model on the paraphrase dataset PPDB, then apply the word representation for word and bi-gram similarity tasks.

Several pre-trained word embeddings are available, which are trained on various corpora under different models. \citet{levy2014dependency} observed that different word embedding models capture different aspects of linguistic properties: a Bag-of-Words contexts based model tends to reflect the domain aspect (e.g., scientist and research) while a paraphrase-relationship based model captures semantic similarities of words (e.g., boy and kid). From experiments, we also observed that the performance of a word embedding model is usually inconsistent over different datasets. This inspired us to develop a model taking advantages of various pre-trained word embeddings for measuring textual similarity/relation.

In this paper, we propose a convolutional neural network (CNN) to learn a multi-aspect word embedding from various pre-trained word embeddings. We then apply the max-pooling scheme and Long Short Term Memory (LSTM) on this embedding to form a sentence representation. In STS tasks, \citet{shao2017hcti} shows the efficiency of the max-pooling scheme in modeling sentences from word embedding representations refined via CNN. However, the max-pooling scheme lacks the property of word order (e.g., \textit{sentence}(``Bob likes Marry'') =  \textit{sentence}(``Marry likes Bob'')). To address this weakness, we use LSTM as an additional scheme for modeling sentences with word order characteristics. For measuring the similarity/relation between two sentence representations, we propose Multi-level comparison which consists of word-word level, sentence-sentence level, and word-sentence level. Through these levels, our model comprehensively evaluates the similarity/relation between two sentences.

We evaluate our M-MaxLSTM-CNN model on three tasks: STS, textual entailment recognition, paraphrase identification. The advantages of M-MaxLSTM-CNN are: i) simple but efficient for combining various pre-trained word embeddings with different dimensions; ii) using Multi-level comparison shows better performances compared to using only sentence-sentence comparison; iii) does not require hand-crafted features (e.g., alignment features, Ngram overlaps, syntactic features, dependency features) compared to the state-of-the-art ECNU \cite{tian2017ecnu} on STS Benchmark dataset.

Our main contributions are as follows:
\begin{itemize}
\item Propose MaxLSTM-CNN encoder for efficiently encoding sentence embeddings from multiple word embeddings.
\item Propose Multi-level comparison (M-MaxLSTM-CNN) to learn the similarity/relation between two sentences. The model achieves strong performances over various tasks. Especially in STS tasks, the method obtains the state-of-the-art results. 
\end{itemize}

The remainder of this paper is organized as follows: Section 2 reviews the previous research, Section 3 introduces the architecture of our model, Section 4 describes the three tasks and datasets, Section 5 describes the experiment setting, Section 6 reports and discusses the results of the experiments, and Section 7 concludes our work. 

\section{Related work}
Most prior research on modeling textual similarity relied on feature engineering. \citet{Wan2006UsingDF} extract $n$-gram overlap features and dependency-based features, while \citet{Madnani} employ features based on machine translation metrics. \citet{mihalcea2006corpus} propose a method using corpus-based and knowledge-based measures of similarity. \citet{das2009paraphrase} design a model which incorporates both syntax and lexical semantics using dependency grammars. \citet{Yangfeng} combine the fine-grained n-gram overlap features with the latent representation from matrix factorization. \citet{Xu} develop a latent variable model which jointly learns paraphrase relations between word and sentence pairs. Using Dependency trees, \citet{Sultan} propose a robust monolingual aligner and successfully applied it for STS tasks.

The recent emergence of deep learning models has provided an efficient way to learn continuous vectors representing words/sentences. By using a neural network in the context of a word prediction task, \citet{bengio2003neural} and \citep{mikolov2013efficient} generate word embedding vectors carrying semantic meanings. The embedding vectors of words which share similar meanings are close to each other. To capture the morphology of words, \citet{Bojanowski} enrich the word embedding with character n-grams information. Closest to this approach, \citet{wieting2016charagram} also propose to represent a word or sentence using a character n-gram count vector. However, the objective function for learning these embeddings is based on paraphrase pairs. 

For modeling sentences, composition approach attracted many studies. \citet{yessenalina2011compositional} model each word as a matrix and used iterated matrix multiplication to present a phrase. \citet{Tai} design a Dependency Tree-Structured LSTM for modeling sentences. This model outperforms the linear chain LSTM in STS tasks. Convolutional neural network (CNN) has recently been applied efficiently for semantic composition \citep{Kalchbrenner,kim2014convolutional,shao2017hcti}. This technique uses convolutional filters to capture local dependencies in term of context windows and applies a pooling layer to extract global features. \citet{he2015multi} use CNN to extract features at multiple level of granularity. The authors then compare their sentence representations via multiple similarity metrics at several granularities. \citet{gan2017learning} propose a hierarchical CNN-LSTM architecture for modeling sentences. In this approach, CNN is used as an encoder to encode an sentence into a continuous representation, and LSTM is used as a decoder. \citet{infersent} train a sentence encoder on a textual entailment recognition database using a BiLSTM-Maxpooling network. This encoder achieves competitive results on a wide range of transfer tasks.  

At SemEval-2017 STS task, hybrid approaches obtain strong performances. \citet{wu2017bit} train a linear regression model with WordNet, alignment features and the word embedding \textit{word2vec}\footnote{https://code.google.com/p/word2vec/}. \citet{tian2017ecnu} develop an ensemble model with multiple boosting techniques (i.e., Random Forest, Gradient Boosting, and XGBoost). This model incorporates traditional features (i.e., n-gram overlaps, syntactic features, alignment features, bag-of-words) and sentence modeling methods (i.e., Averaging Word Vectors, Projecting Averaging Word Vectors, LSTM).

MVCNN model \cite{K15-1021} and MGNC-CNN model \cite{N16-1178}  are close to our approach. In MVCNN, the authors use variable-size convolution filters on various pre-trained word embeddings for extracting features. However, MVCNN requires word embeddings to have the same size. In MGNC-CNN, the authors apply independently CNN on each pre-trained word embedding for extracting features and then concatenate these features for sentence classification. By contrast, our M-MaxLSTM-CNN model jointly applies CNN on all pre-trained word embeddings to learn a multi-aspect word embedding. From this word representation, we encode sentences via the max-pooling and LSTM. To learn the similarity/relation between two sentences, we employ Multi-level comparison.

\begin{figure*}[!ht]
  	\centering
  	\includegraphics[width=320pt]{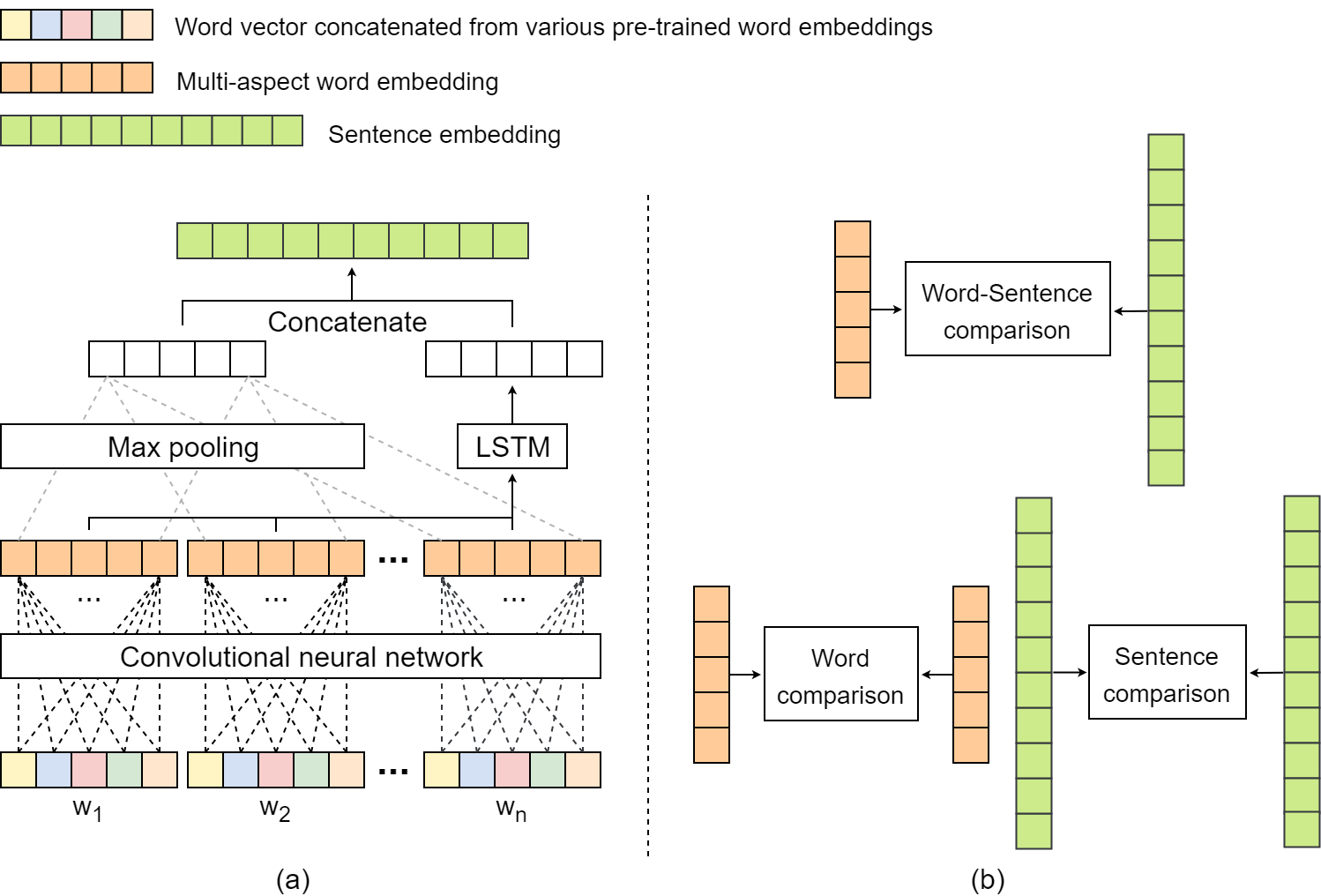}
   	\caption{The proposed M-MaxLSTM-CNN model: (a) MaxLSTM-CNN encoder; (b) Multi-level comparison.}
    \label{fig:model} 
\end{figure*}

\section{Model description}
Our model (shown in Figure \ref{fig:model}) consists of three main components: i) learning a multi-aspect word embedding (Section 3.1); ii) modeling sentences from this embedding (Section 3.2); iii) measuring the similarity/relation between two sentences via Multi-level comparison (section 3.3).

\subsection{Multi-aspect word embedding}
Given a word $w$, we transfer it into a word vector $e^{concat}_w$ via $K$ pre-trained word embeddings as follows:

\begin{equation}
	e^{concat}_w = e^1_w \oplus e^2_w\oplus...\oplus e^K_w
\end{equation}
where $\oplus$ is concatenation operator, $e^i_w$ is the word embedding vector of $w$ in the $i$th pre-trained embedding.

To learn a multi-aspect word embedding $e^{multi}_w$ from the representation $e^{concat}_w$, we design $H$ convolutional filters. Each filter $r_i$ is denoted as a weight vector with the same dimension as $e^{concat}_w$ and a bias value $b_{r_i}$. The $e^{multi}_w$ is obtained by applying these filters on the $e^{concat}_w$ as follows:

\begin{align}
	e^{r_i}_w   &= \sigma(e^{concat}_w r^T_i + b_{r_1})\\
	e^{multi}_w  &= [e^{r_1}_w, e^{r_2}_w, ..., e^{r_H}_w]    
\end{align}
where $\sigma$ denotes a logistic sigmoid function.

The next section explains how to model a sentence from its multiple-aspect word embeddings.

\subsection{Sentence modeling}
Given an input sentence $s=[w_1,w_2,...,w_n]$, we obtain a sequence of multiple-aspect word embeddings $s^{multi}=[e^{multi}_{w1}, e^{multi}_{w2},...,e^{multi}_{w_n}]$ using Eq. (1-3). For modeling the sentence from the representation $s^{multi}$, we use two schemes: max-pooling and LSTM.

\textbf{Max-pooling scheme}: To construct a max-pooling sentence embedding $e^{max}_s$, the most potential features are extracted from the representation $s^{multi}$ as follows:

\begin{equation}
	   e^{max}_s[i] = max(e^{multi}_{w_1}[i], e^{multi}_{w_2}[i], ...,e^{multi}_{w_n}[i])
\end{equation}
where $e^{multi}_{w_k}[i]$ is the $i$th element of $e^{multi}_{w_k}$.

\textbf{LSTM scheme}: From Eq. (4), we find that the max-pooling scheme ignores the property of word order. Therefore, we construct a LSTM sentence embedding $e^{lstm}_s$ to support the sentence embedding $e^{max}_s$. The representation $s^{multi}$ is transformed to a fix-length vector by recursively applying a LSTM unit to each input $e^{multi}_{w_t}$ and the previous step $h_{t-1}$. At each time step $t$, the LSTM unit with $l$-memory dimension defines six vectors in $\mathbb{R}^{l}$: input gate $i_{t}$, forget gate $f_{t}$, output gate $o_{t}$, tanh layer $u_{t}$, memory cell $c_{t}$ and hidden state $h_{t}$ as follows (from \citet{Tai}): 
\begin{align}
	i_{t}&=\sigma(W_{i}e^{multi}_{w_t}+U_{i}h_{t-1}+b_i)\\
    f_{t}&=\sigma(W_{f}e^{multi}_{w_t}+U_{f}h_{t-1}+b_f)\\
    o_{t}&=\sigma(W_{o}e^{multi}_{w_t}+U_{o}h_{t-1}+b_o)\\
    u_t&=\tanh(W_{u}e^{multi}_{w_t}+U_{u}h_{t-1}+b_u)
\end{align}   
\begin{align}
    c_{t}&=f_{t}\odot c_{t-1}+i_{t}\odot u_t\\
    h_{t}&=o_{t}\odot\tanh(c_t)  \\
    e^{lstm}_s& = h_n
\end{align}
where $\sigma, \odot$ respectively denote a logistic sigmoid function and element-wise multiplication; $W_{i}, U_{i}, b_i$ are respectively two weights matrices and a bias vector for input gate $i$. The denotation is similar to forget gate $f$, output gate $o$, tanh layer $u$, memory cell $c$ and hidden state $h$. 

Finally, the sentence embedding $e_s$ is obtained by concatenating the two sentence embeddings $e^{max}_s$ and $e^{lstm}_s$:
\begin{equation}
	e_s = e^{max}_s \oplus e^{lstm}_s
\end{equation}

\subsection{Multi-level comparison}
In this section, we describe the process for evaluating the similarity/relation between two sentences. We compare two sentences via three levels: word-word, sentence-sentence and word-sentence.

\subsubsection{Word-word comparison}
Given two input sentences $s_1$ and $s_2$, we encode them into two sequences of multi-aspect word embeddings $s_1^{multi}$ and $s_2^{multi}$ (Section 3.2). We then compute a word-word similarity vector $sim^{word}$ as follows:
\begin{align}	
    A_{ij} &= \frac{s_1^{multi}[i]\cdot s_2^{multi}[j]}{\left\|s_1^{multi}[i]  \right\| \left\|s_2^{multi}[j]  \right\| } \\
    sim^{word} &= \sigma(W^{word}g(A) + b^{word})
\end{align}
where $s_t^{multi}[i]$ is the $i$th multi-aspect word embedding of sentence $s_t$; $g()$ is a function to flatten a matrix into a vector; and $W^{word}$ and $b^{word}$ are respectively a weight matrix and a bias parameter.

\subsubsection{Sentence-sentence comparison}
Given two input sentences $s_1$ and $s_2$, we encode them into two sentence embeddings $e_{s_1}$ and $e_{s_2}$ (Section 3.1 and 3.2). To compute the similarity/relation between the two embeddings, we introduce four comparison metrics:

\textbf{Cosine similarity}:
\begin{equation}
	d_{cosine} = \frac{e_{s_1}\cdot e_{s_1}}{\left\|e_{s_1} \right\|\left\|e_{s_2} \right\|} 
\end{equation}

\textbf{Multiplication vector \& Absolute difference}:  
\begin{align}	
    d_{mul} &= e_{s_1} \odot e_{s_2}  \\
    d_{abs}& = |e_{s_1} - e_{s_2}|
\end{align}
where $\odot$ is element-wise multiplication.

\textbf{Neural difference}:
\begin{align}	
    x &= e_{s_1} \oplus e_{s_2}  \\
    d_{neu} &= W^{neu}x + b^{neu}
\end{align}
where $W^{neu}$ and $b^{neu}$ are respectively a weight matrix and a bias parameter.

As a result, we have a sentence-sentence similarity vector $sim^{sent}$ as follows:
\begin{align}
	d^{sent} &=  d_{cosine} \oplus d_{mul} \oplus d_{abs} \oplus d_{neu}\\ 
    sim^{sent} &= \sigma(W^{sent}d^{sent} + b^{sent})
\end{align}
where $W^{sent}$ and $b^{sent}$ are respectively a weight matrix and a bias parameter.

\subsubsection{Word-sentence comparison}
Given a sentence embedding $e_{s_1}$ and a sequence of multi-aspect word embeddings $s_2^{multi}$, we compute a word-sentence similarity matrix $sim_{s_1}^{ws}$ as follows:
\begin{align}	
	e_{s_1}^{ws}[i] &= e_{s_1} \oplus s_2^{multi}[i]\\
    sim_{s_1}^{ws}[i] &= \sigma(W^{ws}e_{s_1}^{ws}[i] + b^{ws})
\end{align}
where $s_2^{multi}[i]$ is the multi-aspect word embedding of the $i$th word in sentence $s_2$; $W^{ws}$ and $b^{ws}$ are respectively a weight matrix and a bias parameter.

As a result, we have a word-sentence similarity vector $sim^{ws}$ for the two sentences as follows:
\begin{equation}
	sim^{ws} = \sigma(W^{ws'}[g(sim_{s_1}^{ws}) \oplus g(sim_{s_2}^{ws})]+b^{ws'})
\end{equation}
where $g()$ is a function to flatten a matrix into a vector; $W^{ws'}$ and $b^{ws'}$ are respectively a weight matrix and a bias parameter.

Finally, we compute a target score/label of a sentence pair as follows:
\begin{align}	
    sim &= sim^{word} \oplus sim^{sent} \oplus sim^{ws}\\
 	h_s &= \sigma(W^{l1}sim+b^{l1})\\
    \hat{y} &= softmax(W^{l2}h_s+b^{l2})
\end{align}
where $W^{l1}$, $W^{l2}$, $b^{l1}$ and $b^{l2}$ are model parameters; $\hat{y}$ is a predicted target score/label.

\section{Tasks \& Datasets}
We evaluate our model on three tasks:
\begin{itemize}
\item \textbf{Textual entailment recognition}:
given a pair of sentences, we predict a directional relation between the sentences (entailment/contradiction/neutral). We evaluate this task on \textbf{SICK} dataset. It was collected for the 2014 SemEval competition and includes examples of the lexical, syntactic and semantic phenomena and ignores other aspects of existing sentential datasets (i.e., idiomatic multiword expressions, named entities, telegraphic language).

\item  \textbf{Semantic textual similarity}:
given a pair of sentences, we measure a semantic similarity score of this pair. We use two datasets for this task:
\begin{itemize}
\item \textbf{STSB}: comprises a careful selection of the English data sets used in SemEval and *SEM STS shared tasks from 2012 to 2017. This dataset cover three genres: image captions, news headlines and user forums. Each sentence pair is annotated with a relatedness score $\in[0,5]$.

\item \textbf{SICK}: Each sentence pair is annotated with a relatedness score $\in[1,5]$.

\end{itemize}

\item \textbf{Paraphrase identification}:
given a pair of sentences, we predict a binary label indicating whether the two sentences are paraphrases. Microsoft Research Paraphrase Corpus (MRPC) is used for this task. It includes pairs of sentences extracted from news source on the web.

\end{itemize}

Table \ref{tab:dataset} shows the statistic of the three datasets. Because of not dealing with name entities and multi-word idioms, the vocabulary size of SICK is quite small compared to the others.

\begin{table}[!ht]
\small
\centering
\setlength{\tabcolsep}{6pt}
\renewcommand{\arraystretch}{1.2}
\begin{tabular}{l | c c c  c c}
\hline
\multicolumn{1}{c|}{\textbf{Dataset}} & \textbf{Train}&\textbf{Validation}&\textbf{Test} & $l$ & $|V|$\\ 

\hline
STSB & 5,749&1,500&1,379 & 11& 15,997\\
SICK         & 4,500&500&4,927 & 9 & 2,312\\
MRPC         & 3,576&500&1,725 & 21 & 18,003\\
\hline
\end{tabular}

\caption{Statistic of datasets. $|V|$, $l$ denote the vocabulary size, and the average length of sentences respectively. }
\label{tab:dataset}
\end{table}

\section{Experiment setting}
\subsection{Pre-trained word embeddings}
We study five pre-trained word embeddings\footnote{These embeddings are available at \textit{anonymous}} for our model:
\begin{itemize}
\item \textbf{word2vec} is trained on Google News dataset (100 billion tokens). The model contains 300-dimensional vectors for 3 million words and phrases.

\item \textbf{fastText} is learned via skip-gram with subword information on Wikipedia text. The embedding representations in fastText are 300-dimensional vectors.

\item \textbf{GloVe} is a 300-dimensional word embedding model learned on aggregated global word-word co-occurrence statistics from Common Crawl (840 billion tokens).

\item \textbf{Baroni} uses a context-predict approach to learn a 400-dimensional semantic embedding model. It is trained on 2.8 billion tokens constructed from ukWaC, the English Wikipedia and the British National Corpus. 

\item \textbf{SL999} is trained under the skip-gram objective with negative sampling on word pairs from the paraphrase database \textbf{PPDB}. This 300-dimensional embedding model is tuned on SimLex-999 dataset \citep{hill2016simlex}.  

\end{itemize}
\subsection{Model configuration}
In all of the tasks, we used the same model configuration as follows:
\begin{itemize}
\item Convolutional filters: we used 1600 filters. It is also the dimension of the word embedding concatenated from the five pre-trained word embeddings.
\item LSTM dimension: we also selected 1600 for LSTM dimension.
\item Neural similarity layers: the dimension of $b^{word}$, $b^{sent}$,  $b^{ws}$ and $b^{ws'}$ are respectively 50, 5, 5 and 100.
\item Penultimate fully-connected layer: has the dimension of 250 and is followed by a drop-out layer ($p = 0.5$). 

\end{itemize}

We conducted a grid search on 30\% of STSB dataset to select these optimal hyper-parameters.
\subsection{Training Setting}
\subsubsection{Textual entailment recognition \& Paraphrase identification}

In these tasks, we use the cross-entropy objective function and employ AdaDelta as the stochastic gradient descent (SGD) update rule with mini-batch size as 30. Details of Adadelta method can be found in \citet{zeiler2012adadelta}. During the training phase, the pre-trained word embeddings are fixed. 

\subsubsection{Semantic Textual Similarity}
To compute a similarity score of a sentence pair in the range $[1,K]$, where $K$ is an integer, we replace Eq. (27) with the equations in \citet{Tai} as follows:
\begin{align}
    \hat{p}_\theta &= softmax(W^{l2}h_s+b^{l2})\\
    \hat{y} &= r^T\hat{p}_\theta
\end{align}
where $W^{l1}$, $W^{l2}$, $b^{l1}$ and $b^{l2}$ are parameters; $r^T = [1,2,...,K]$; $\hat{y}$ is a predicted similarity score.

A sparse target distribution $p$ which satisfies $y=r^Tp$ is computed as:

\begin{equation}
	p_i = \left\{\begin{matrix}
y- \left \lfloor y \right \rfloor , & i =\left \lfloor y \right \rfloor + 1\\ 
\left \lfloor y \right \rfloor  - y + 1, & i=\left \lfloor y \right \rfloor \\ 
0 & otherwise
\end{matrix}\right.
\end{equation}
for $i\in[1,K]$, and $y$ is the similarity score.

To train the model, we minimize the regularized KL-divergence between $p$ and $\hat{p}_\theta$:
\begin{equation}
J(\theta)=\frac{1}{m}\sum^m_{k=1}KL(p^{(k)}||\hat{p}^{(k)}_\theta)
\end{equation}
where $m$ is the number of training pairs and $\theta$ denotes the model parameters. The gradient descent optimization Adadelta is used to learn the model parameters. We also use mini-batch size as 30 and keep the pre-trained word embeddings fixed during the training phase. We evaluate our models through Pearson correlation $r$.

\begin{table*}[!ht]
\centering
\small
\renewcommand{\arraystretch}{1.1}
\setlength{\tabcolsep}{10pt}
\begin{tabular}{  p{70mm} | c c c c}
\hline
\multicolumn{1}{c|}{\textbf{Method}} & \textbf{STSB} & \textbf{SICK-R} & \textbf{SICK-E} & \textbf{MRPC}   \\\hline

\multicolumn{5}{l}{\textit{Ensemble models/Feature engineering}} \\\hline
DT\_TEAM \citep{maharjan2017dt_team} &79.2 &- & - & -\\
ECNU \citep{tian2017ecnu} & 81 & - & - & -\\
BIT \citep{wu2017bit} & 80.9 & - & - & -\\
TF-KLD \citep{Yangfeng} & -&- & -&\textbf{80.41}/\textbf{85.96}\\
\hline
\multicolumn{5}{l}{\textit{Neural representation models with one embedding}} \\\hline
Multi-Perspective CNN \citep{he2015multi} & - & 86.86 & - & 78.6/84.73\\
InferSent \citep{infersent} & 75.8 & 88.4 & \textbf{86.1} & 76.2/83.1\\
GRAN \citep{gran}& 76.4 & 86 & - &-\\
Paragram-Phrase \citep{wieting2015towards} &73.2 & 86.84 & 85.3 & -\\
HCTI \citep{shao2017hcti} & 78.4 & -  & - &-\\\hline
\multicolumn{5}{l}{\textit{Neural representation models with the five embeddings using sentence-sentence comparison (S)}} \\\hline
S-Word Average  & 71.06 & 81.18   & 80.88  & 71.48/81.1\\
S-Project Average  & 75.12 & 86.53  & 85.12 & 75.48/82.47\\
S-LSTM & 77.14 & 85.15  & 85.6 & 70.43/79.71\\
S-Max-CNN & 81.87 & 88.3  & 84.33 & 76.35/83.75\\
S-MaxLSTM-CNN & 82.2 & 88.47   & 84.9 &77.91/84.31\\\hline

\multicolumn{5}{l}{\textit{Neural representation models with the five embeddings using Multi-level comparison (M)}} \\\hline
M-Max-CNN & 82.11 & 88.45  & 84.7 & 76.75/83.64\\
M-MaxLSTM-CNN & \textbf{82.45} & \textbf{88.76}   & 84.95 &78.1/84.5\\

\hline
\end{tabular}
\caption{Test set results with Pearson's $r$ score$\times 100$ for STS tasks, and accuracy for other tasks. Boldface values show the highest scores in each dataset. SICK-R and SICK-E denote the STS task and the entailment task in SICK dataset respectively.}\label{tab:overall}
\end{table*}

\section{Experiments and Discussion}
This section describes two experiments: i) compare our model against recent systems; ii) evaluate the efficiency of using multiple pre-trained word embeddings.

\subsection{Overall evaluation}
Besides existing methods, we also compare our model with several sentence modeling approaches using multiple pre-trained word embeddings: 

\begin{itemize}
\item \textbf{Word Average}:
\begin{equation}
e_s = \frac{1}{n}\sum^n_{i=1}{e^{concat}_{w_i}}
\end{equation}
where $e_s$ is the sentence embedding of a $n$-words sentence, and $e^{concat}_{w_i}$ is from Eq. (1)

\item \textbf{Project Average}:
\begin{equation}
e_s = \sigma(W(\frac{1}{n}\sum^n_{i=1}{e^{concat}_{w_i}}) + b)
\end{equation}
where $W$ is a $1600\times1600$ weight matrix, and $b$ is a $1600$ bias vector.

\item \textbf{LSTM}: apply Eq. (5-11) on $e^{concat}_{w_i}$ to construct the $1600$-dimension $e_s$ sentence embedding. 

\item\textbf{Max-CNN}: apply Eq. (2-4) on $e^{concat}_{w_i}$ to construct the $1600$-dimension $e_s$ sentence embedding. 
\end{itemize}

We report the results of these methods in Table \ref{tab:overall}. Overall, our M-MaxLSTM-CNN shows competitive performances in these tasks. Especially in the STS task, M-MaxLSTM-CNN outperforms the state-of-the-art methods on the two datasets. Because STSB includes complicated samples compared to SICK, the performances of methods on STSB are quite lower. In STSB, the prior top performance methods use ensemble approaches mixing hand-crafted features (word alignment, syntactic features, N-gram overlaps) and neural sentence representations, while our approach is only based on a neural sentence modeling architecture. In addition, we observed that InferSent shows the strong performance on SICK-R but quite low on STSB while our model consistently obtains the strong performances on both of the datasets. InferSent uses transfer knowledge on textual entailment data, consequently it obtains the strong performance on this entailment task.

According to \citet{wieting2015towards}, using Word Average as the compositional architecture outperforms the other architectures (e.g., Project Average, LSTM) for STS tasks. In a multiple word embeddings setting, however, Word Average does not show its efficiency. Each word embedding model has its own architecture as well as objective function. These factors makes the vector spaces of word embeddings are different. Therefore, we intuitively need a step to learn or refine a representation from a set of pre-trained word embeddings rather than only averaging them. Because Project Average model, LSTM model and Max-CNN model have their parameters for learning sentence embeddings, they significantly outperform Word Average model.

\begin{table*}[!ht]
\small
\centering
\setlength{\tabcolsep}{8pt}
\renewcommand{\arraystretch}{1.15}
\begin{tabular}{l | c c| c c c | c c}

\multirow{2}{*}{\textbf{Word embedding}} & \multicolumn{2}{c|}{\textbf{STSB}} & \multicolumn{3}{c|}{\textbf{SICK-R} \& \textbf{SICK-E}}& \multicolumn{2}{c}{\textbf{MRPC}}\\ 

\cline{2-8}
 & Pearson& $|V|_{avai}(\%)$&Pearson & Acc & $|V|_{avai}(\%)$ & Acc/F1 & $|V|_{avai}(\%)$ \\ 
\hline
word2Vec& 78.9 & 75.64 & 87.27 &84.09&98.53 & 75.42/82.13  &67.81 \\ 
fastText& 79.95 & 84.27 & 87.59 &83.42&99.18 &   74.31/81.75&79.04\\ 
Glove& 80.1 & 91.71 & 88.21 &84.71&99.78 &  74.9/82.782&89.85\\ 
SL999& 80.31 & 94.76 & 87.26 &84.55&99.83 &76.46/83.13&94.19\\ 
Baroni& 79.81 &90.54  & 86.9 &83.99&98.83&  74.84/82.4&87.92 \\ 
\hline
All & \textbf{82.45} & 95.65 & \textbf{88.76} &\textbf{84.95}&99.83& \textbf{78.1}/\textbf{84.5} &  95.97\\ 
\hline
\end{tabular}

\caption{Evaluation of exploiting multiple pre-trained word embeddings. $|V|_{avai}$ is the proportion of vocabulary available in a word embedding. In case of using all word embeddings, $|V|_{avai}$ denotes the proportion of vocabulary where each word is available in at least one word embedding. }
\label{tab:compemb}
\end{table*}

We observed that MaxLSTM-CNN outperforms Max-CNN in both of the settings (i.e., sentence-sentence comparison, Multi-level comparison). As mentioned in Section 1, Max-CNN ignores the property of word order. Therefore, our model achieves improvement compared to Max-CNN by additionally employing LSTM for capturing this property.

We only applied Multi-level comparison on Max-CNN and MaxLSTM-CNN because these encoders generate multi-aspect word embeddings. The experimental results prove the efficiency of using Multi-level comparison. In the textual entailment dataset \textbf{SICK-E}, the task mainly focuses on interpreting the meaning of a whole sentence pair rather than comparing word by word. Therefore, the performance of Multi-level comparison is quite similar to sentence-sentence comparison in the SICK-E task. This is also the reason why LSTM, which captures global relationships in sentences, has the strong performance in this task.

\subsection{Evaluation of exploiting multiple pre-trained word embeddings}
In this section, we evaluate the efficiency of using multiple pre-trained word embeddings. We compare our multiple pre-trained word embeddings model against models using only one pre-trained word embedding. The same objective function and Multi-level comparison are applied for these models. In case of using one pre-trained word embedding, the dimension of LSTM and the number of convolutional filters are set to the length of the corresponding word embedding. Table \ref{tab:compemb} shows the experimental results of this comparison. Because the approach using five word embeddings outperforms the approaches using two, three, or four word embeddings, we only report the performance of using five word embeddings. We also report $|V|_{avai}$ which is the proportion of vocabulary available in a pre-trained word embedding. SICK dataset ignores idiomatic multi-word expressions, and named entities, consequently the $|V|_{avai}$ of SICK is quite high.

We observed that no word embedding has strong results on all the tasks. Although trained on the paraphrase database and having the highest $|V|_{avai}$, the SL999 embedding could not outperform the Glove embedding in SICK-R. HCTI \citep{shao2017hcti}, which is the current state-of-the-art in the group of neural representation models on STSB, also used the Glove embedding. However, the performance of HTCI in STSB ($78.4$) is lower than our model using the Glove embedding. In SICK-R, InferSent \citep{infersent} achieves a strong performance ($88.4$) using the Glove embedding with transfer knowledge, while our model with only the Glove embedding achieves a performance close to the performance of InferSent. These results confirm the efficiency of Multi-level comparison.

In STSB and MRPC, as employing the five pre-trained embeddings, the $|V|_{avai}$ is increased. This factor limits the number of random values when initializing word embedding representations because a word out of a pre-trained word embedding is assigned a random word embedding representation. In other words, a word out of a pre-trained word embedding is assigned a random semantic meaning. Therefore, the increase of the $|V|_{avai}$ improves the performance of measuring textual similarity. In STSB and MRPC, our multiple pre-trained word embedding achieves a significant improvement in performance compared against using one word embedding. In SICK-R and SICK-E, although the $|V|_{avai}$ is not increased when employing five pre-trained embeddings, the performance of our model is improved. This fact shows that our model learned an efficient word embedding via these pre-trained word embeddings.

\section{Conclusion}
In this work, we study an approach employing multiple pre-trained word embeddings and Multi-level comparison for measuring semantic textual relation. The proposed M-MaxLSTM-CNN architecture consistently obtains strong performances on several tasks. Compared to the state-of-the art methods in STS tasks, our model does not require handcrafted features (e.g., word alignment, syntactic features) as well as transfer learning knowledge. In addition, it allows using several pre-trained word embeddings with different dimensions.

Future work could apply our multiple word embeddings approach for transfer learning tasks. This strategy allows making use of pre-trained word embeddings as well as available resources.

\section*{Acknowledgments}
This work was done while Nguyen Tien Huy was an intern at Toshiba Research Center. 

\bibliography{acl2018}
\bibliographystyle{acl_natbib}

\end{document}